# SDRNET: Stacked Deep Residual Network For Accurate Semantic Segmentation Of Fine-Resolution Remotely Sensed Images

Naftaly Wambugu[a], Ruisheng Wang[a]*, Bo Guo[a], Tianshu Yu[b], Sheng Xu[c], Mohammed Elhassan[d]

[a] *State Key Laboratory of Subtropical Building and Urban Science, School of Architecture and Urban Planning, Shenzhen University, Shenzhen, China*
[b] *The Chinese University of Hong Kong, Shenzhen, China*
[c] *College of Information Science and Technology & Artificial Intelligence, Nanjing Forestry University, Nanjing 210037, China*
[d] *School of Computer Science, Zhejiang Normal University, Jinhua 321004, Zhejiang, China.*

Email: wambugunaph@gmail.com, *ruiswang@szu.edu.cn, guobo@szu.edu.cn, yutianshu@cuhk.edu.cn, xusheng@njfu.edu.cn, mohammedac29@zjnu.edu.cn

**Abstract**
Land cover maps generated from semantic segmentation of high-resolution remotely sensed images have drawn much attention in the photogrammetry and remote sensing research community. Currently, massive fine-resolution remotely sensed (FRRS) images acquired by improving sensing and imaging technologies become available. However, accurate semantic segmentation of such FRRS images is greatly affected by substantial class disparities, the invisibility of key ground objects due to occlusion, and object size variation. Despite the extraordinary potential for deep convolutional neural networks (DCNNs) in image feature learning and representation, extracting sufficient features from FRRS images for accurate semantic segmentation is still challenging. These challenges demand the deep learning models to learn robust features and generate sufficient feature descriptors. Specifically, learning multi-contextual features to guarantee adequate coverage of varied object sizes from the ground scene and harnessing global-local contexts to overcome class disparities challenge even profound networks. Deeper networks significantly lose spatial details due to gradual downsampling processes resulting in poor segmentation results and coarse boundaries. This article presents a stacked deep residual network (SDRNet) for semantic segmentation from FRRS images. The proposed framework uses two stacked encoder-decoder networks to harness long-range semantics yet preserve spatial information. In addition, we introduce dilated residual blocks (DRB) between each encoder and decoder network to capture sufficient global dependencies. The attention blocks adopted in SDRNet refine the learnable features and help improve segmentation performance. Finally, an intermediate loss is introduced mid-way to the network to supervise the learning process at the middle layers. Our experimental results obtained using the ISPRS Vaihingen and Potsdam datasets demonstrated that the SDRNet outperforms some of the current best methods in semantic segmentation by attaining overall accuracy of 90.82% and 90.62%, respectively.

⎕*Corresponding authors at: Shenzhen University, China*
*E-mail: wambugunaph@gmail.com (Wambugu Nafatly)*

# 1. Introduction

Human interest in understanding Earth's surface changes and anthropogenic activities has spurred massive spatial data generation, thanks to the advanced remote sensing data acquisition technologies. Moreover, increased and accessible computational resources and improved algorithms have greatly encouraged the research community to delve more into analysis and in-depth understanding of remotely sensed data with refined precision and detail (Hermosilla et al., 2022).

Semantic segmentation assigns similar pixels in an image to defined classes using an array of feature interpretation approaches. Using automated computational models, the semantic segmentation and classification of fine-resolution remotely sensed (FRRS) images has been a major means for the production of detailed land cover maps.

Despite the derived benefits from RS image analysis, tackling land cover mapping using FRRS images poses great difficulties. The high intra-class disparities among objects of the same class and high inter-class similarities among objects of different classes complicate the classification task (Huang and Zhang, 2013). Moreover, the rich information and spatial detail in FRRS images demand an elaborate automated feature learning scheme. Further, arbitrary object orientation and sizes where large objects suppress smaller objects and occlusion in fine resolution images pose significant challenges for accurate semantic segmentation and make delineating specific class objects difficult (Zhao et al., 2017b).

Traditional approaches that rely on human expertise to define learnable hand-crafted, low-level features cannot sufficiently discriminate complex features in VHR data. They fail to harness the rich contextual information in fine-grained images (Li et al., 2021b). Moreover, these shallow classifiers tend to be tedious, less effective, slow, and domain expert-dependent in image analysis tasks. They are also deemed task-specific and data-dependent low-level feature extractors that are poor in generalization. They may work well for specific cases and fail to perform in others (Volpi and Tuia, 2017).

Generating accurate segmentation maps from remotely sensed data with rich and heterogeneous information demands complex and deep network structures. Convolutional neural networks (CNNs) offer a great paradigm where complex image features are hierarchically and automatically learnt using alternating convolutional layers, sub-sampling layers, and activation functions. CNNs have demonstrated remarkable power in learning complex relationships within high-resolution spatial data, significantly improving feature learning and image analysis in fine-grained images (Pan et al., 2020; Zheng et al., 2020). Owing to their benefits, CNNs have overtaken feature engineering methods. Moreover, they can easily be implemented using existing computational resources such as graphics processing power (GPUs) and distributed computing platforms. Deep convolutional neural networks (DCCNs) are the advanced version of CNNs with deeper layers designed to learn high semantic representation from voluminous data. Recently, DCCNs' superior performance in image-related tasks such as object recognition, object detection, and semantic segmentation and their ability to handle complex and big data has made them gain undisputed popularity over shallow models in image analysis tasks (Dong et al., 2021; Han et al., 2017).

In the general structure of CNNs, multiple convolutional layers learn abstract image representation from the input image at different levels and spatial scales. The sub-sampling layer gradually decreases the images' spatial details through pooling operations that capture a larger field of view and reduce the number of learnable parameters. However, as the network deepens to harness high-level image features and complex representation from images, essential spatial details suffer critically due to multiple down-sampling operations in CNNs, resulting in coarse feature maps. Thus, spatial details are indispensable for accurate semantic segmentation, especially at the pixel level (Zhang et al., 2020a). Arguably, deeper networks in practice tend to be more potent in dealing with non-linear and complex relationships between data. Moreover, when extracting high-level and detailed structural image features such as complex RS data, deeper networks have been preferred over shallow counterparts (Diakogiannis et al., 2020; Huang et al., 2018; Rezaee et al., 2018).

Since deeper networks can result in vanishing gradient problems that harm network performance, residual learning (He et al., 2016) presented a viable and straightforward method of improving gradient flow, thus enabling the design of deeper networks for image analysis tasks. Consequently, following the residual learning concept, very deep CNNs have been developed for image analysis tasks such as image recognition (Wu et al., 2019), image classification (Wang et al., 2017; Zhang et al., 2020b), semantic segmentation (Diakogiannis et al., 2020; Samy et al., 2018), among others. Thus, residual learning proves to be a promising intuitive method that enables the design and implementation of very deep convolutional networks.

The successful application of CNN-based models for semantic segmentation tasks can be traced back to the novel concept of fully convolutional neural network (FCN) (Long et al., 2015), which performed pixel-level labeling of high-resolution aerial imagery in an end-to-end fashion. Later, following the FCN concept, several other improved networks have been proposed for image analysis tasks, such as high-resolution image classification (Fu et al., 2017), road extraction (Buslaev et al., 2018b), land cover mapping (Mboga et al., 2020), among others. However, FCN-based networks suffer from loss of spatial information due to sub-sampling operations such as pooling and strided convolutions, thus cannot work well with fine-grained images. Volpi and Tuia (2017). improved FCN architecture by proposing transposed convolution in a down-sampling-up-sampling scheme to cure spatial details loss. Later, elaborate networks were developed following the down-sampling and upsampling approach. For instance, UNet (Ronneberger et al., 2015), initially designed for medical image classification, uses two symmetrical paths. The analysis phase, where image semantic details are extracted, and the synthesis phase, where spatial and positional details are added to the deeper layers to refine the segmentation output (Li et al., 2021a; Ronneberger et al., 2015). In another attempt to solve the loss of spatial details, Badrinarayanan et al. (2017) proposed SegNet for image segmentation that uses pooling indices on the decoder to perform the upsampling operation, thus raising the spatial details.

Another challenge with segmentation models is harnessing both global and local information. Models designed for segmentation tasks perform poorly when only local information is available to complement the images' local information with the global context. Following the PsPNet (Zhao et al., 2017a) paradigm, fusing global and local contextual information improved segmentation accuracy since multi-scale semantic features were utilized comprehensively to delineate semantic classes for varying spatial scales. Later, Deeplab and its variants (Chen et al., 2018a; Chen et al., 2018b) combined depth-wise separable convolution and atrous spatial pyramid pooling to learn images' multi-contextual information, where DeepLabV3+ employs a decoder subnetwork to refine the segmentation outputs. These methods have revolutionized the semantic segmentation tasks and posted increasingly improved accuracy but still cannot adequately handle semantic segmentation using complex FRRS images. Specifically, obtaining sufficient multi-contextual information and harnessing sufficient spatial details is difficult for semantic segmentation of FRRS images. Therefore, designing a robust network model for remote sensing data semantic segmentation presents considerable difficulties.

Motivated by our previous work (Wambugu et al., 2021), in this paper, we propose a more computationally efficient framework that improves the previous CRD-Net model to handle land cover classification. The SDRNet framework presents an effective multi-level feature extraction scheme that extracts multi-level features from the network backbone to improve the segmentation performance. The network employs stacked subnetworks to intensify the learning of complex image features while preserving the essential location details using an encoder-decoder approach. The SDRNet employs an improved encoder with ResNet50 with four blocks at both the first and the second encoder to reduce the number of parameters. Moreover, dilated residual blocks are implemented between each encoder and decoder to raise the receptive field, harness multi-contextual features hierarchically, and obtain a more detailed segmentation effect.

Unlike the CRD module used in the CRD-Net that uses parallel input connections,

the DRB blocks have the same structural design but use serial connections between the dilated layers. In addition, the attention blocks introduced at each of the encoder segments refine the semantic features. The SDRNet framework employs extensive residual connections between the encoder and decoder sub-networks to promote elaborate information flow, elevating gradient degradation prone to very deep networks.

The contributions of our article are as follows:

1. We propose a novel SDRNet to tackle the challenge in semantic segmentation of FRRS images.

2. We propose a stacked encoder-decoder sub-networks to advance multi-level feature learning. The dilated residual blocks enlarge the receptive field to harness multi-scale inference and attention mechanism to refine the essential learnable features.

3. The SDRNet framework is validated on the ISPRS Potsdam and Vaihingen datasets and attains competitive classification results in the two datasets without post-processing consideration.

The rest of this paper is organized as follows. Section 2 reviews related work. Section 3 details our method. Section 4 presents and discusses our experimental results. Section 5 concludes the paper.

**2. Related work**

This section reviews some of the recent works in the literature relevant to our work in semantic segmentation for VHR remotely sensed imagery.

Semantic segmentation of VHR images poses great challenges due to the complex nature of the RS images, yet, its importance and value cannot be underestimated. The varying object sizes, arbitrary orientations, and shapes make the semantic segmentation tasks of RS images difficult. Moreover, when dealing with land cover data, same class objects can exhibit different structural and visual characteristics, where various land cover objects can exhibit complex features and visual aspects (large intra-class disparities and low inter-class variance) (Pan et al., 2018). When dealing with FRRS images, the situation worsens, leading to incoherent semantic segmentation results.

Most models utilize image multiscale semantics to handle varied sized objects and multi-object distribution. Dilated convolution (Yu and Koltun, 2016) obtains multiscale features by expanding the receptive field by introducing holes into the regular convolution while maintaining the feature 'map's resolution (max-pooling or strided convolution in standard convolution reduces the feature map resolution harming segmentation accuracy). The dilation convolution has a hyperparameter called dilation rate, which refers to the number of intervals of the convolution kernel compared to the original standard convolution. The main advantage of dilated convolution is allowing flexible adjustment of the 'filter's receptive field to capture multiscale information without loss of spatial information or increasing parameters resulting in the accurate delineation of semantic boundaries (Liu et al., 2020).

Although dilated convolution is preferred in several semantic segmentation tasks for alleviating the discrepancy between the resolution of the feature map and the size of the receptive field, it still presents significant challenges (Gomes et al., 2021). For example, since the neurons in the feature map have the same receptive field, the semantic mask generation process uses features at only one scale (Chen et al., 2018b). When the dilation rate increases, it causes gridding problem or "the checkerboard effect" due to lack of correlation between the adjacent features. Gridding effects make the adoption of large dilated kernels in dilated convolutions practically difficult.

Several networks have been proposed following encoder-decoder and multi-contextual feature aggregation methods for semantic segmentation to handle diminished feature resolution to solve the issue of loss of spatial information. ScasNet (Liu et al., 2018) used a CNN encoder to capture global and local contextual information and later employed a coarse-to-fine refinement scheme. ShelfNet (Zhuang et al., 2019) used multiple encoder-decoder branch pairs with skip connections at different spatial levels to reconstruct location details. ResUNet-a (Diakogiannis et al., 2020) employed an encoder-decoder UNet backbone coupled with a multi-contextual aggregation scheme that uses pyramid pooling and atrous convolution to perform semantic segmentation on FRRS data. In other works, HCANet (Bai et al., 2022) adopted the encoder-decoder structure with a compact atrous pyramid pooling embedded on the decoder for strong aggregation of multi contextual information to deal with the classification of VHR-RS images.

Most semantic segmentation models fail to harness global and local information in segmentation tasks sufficiently. The model performs poorly when only local information is available, hence the need to complement the images' local information with the global context. For example, MANet (Li et al., 2022b) used multiple attention modules to extract contextual dependencies and improve local and global dependencies to improve semantic segmentation results. SSAtNet (Zhao et al., 2022) proposed an end-to-end attention-based network that uses a pyramid attention pooling module and attention mechanism multiscale feature learning and adaptive features refinement. In contrast, AFNet (Yang et al., 2021) uses a multipath encoder structure to extract features of multipath inputs by combining a multipath attention-fused block to fuse multipath features. Moreover, a refinement attention-fused block module fuses high-level abstract features and low-level spatial features. These works show the importance of contextual feature aggregation using multiscale feature learning and feature refinement.

From these studies, we observe that a lot of effort has been made toward improving the semantic segmentation of fine-grained remote sensing imagery using DCNN models. Most of the above works demonstrate that dilated convolution, encoder-decoder paradigm, and attention-based methods are preferred for harnessing multi-contextual features, preserving spatial details for better semantic segmentation results, and feature refinement. Still, some works combine two or more approaches to develop a robust approach resulting in improved segmentation results. However, developing a robust and efficient CNN-based model for semantic segmentation using complex FRRS images still presents significant difficulties, and existing methods require improvement.

## 3. Our method

This section details the architecture of SDRNet. Deep feature extraction using stacked encoder-decoder network is explained in Section 3.2, while multi-feature aggregation using DRB blocks is covered in section 3.3. Lastly, section 3.4 highlights training supervision using intermediate loss.

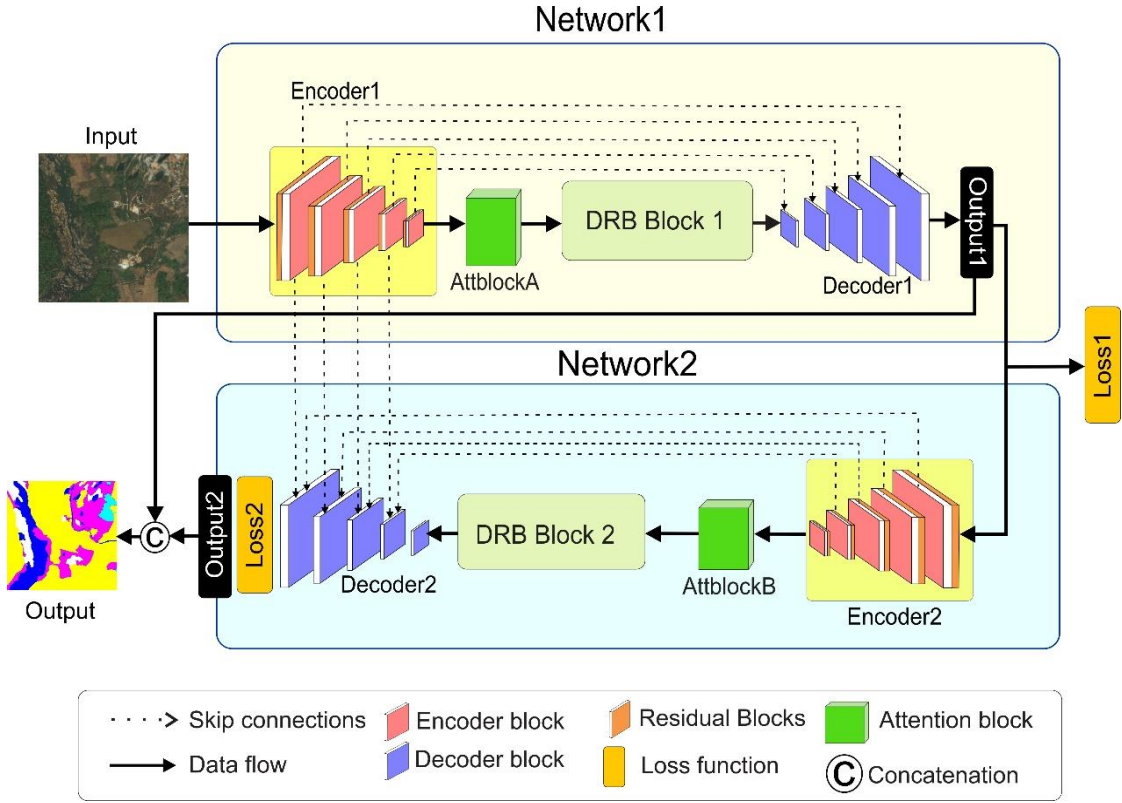

**Fig. 1.** The proposed stacked deep residual network (SDRNet)

### 3.1 Network architecture

We propose the SDRNet framework to handle land cover classification tasks using fine resolution remotely sensed images. A more detailed description of our proposed framework is provided in later sections. The overall SDRNet architecture is presented

in **Fig. 1.**

The structure of SDRNet is made of two subnetworks hierarchically stacked together. Each subnetwork comprises an encoder, spatial attention blocks, the dilated residual module, and the decoder. We use the encoder-decoder concept because pixel-wise classification largely depends on full RF to capture long-range semantics from the input image. Most of the RS scenes have varying object sizes, and thus the requirement for larger RF is inevitable. Also, since small-sized ground objects leverage spatial details for correct segmentation, the two decoders significantly support the reconstruction of the spatial information lost in the initial and lower layers of the encoding phase. Our proposed framework also leverages the network's backbone to harness multi-level feature representative for valid classification results yet effectively handles the vanishing gradient problem prone to deep networks. In the following subsections, we describe the proposed SDRNet in detail.

*3.2 Deep feature extraction using stacked encoder-decoder network*

It is challenging to design a robust network that can capture all the necessary features for accurate land cover classification using FRRS data. CNNs have proven powerful and effective in learning complex and abstract image representation. However, the down-sampling procedures of pooling cause loss of spatial information considered indispensable for pixel-level classification. It gets more profound when dealing with FRRS data that often have variable-sized objects, with cases of large objects suppressing small-sized objects. Therefore, accurate segmentation demands that the network preserves the essential spatial details.

Our proposed framework employs the encoder-decoder paradigm to lower image spatial dimensionality and, at the same time, increase the receptive field. The earliest layers in the network learn low-level image features such as edges, blobs, and orientations, while the subsequent layers learn more distinguishable image features. Reducing the image's spatial dimensions allows the network to focus on high-level semantics as the network deepens in the encoding phase. Large objects from the RS scene can be captured by increasing the receptive field. As shown in **Fig.2,** Encoder 1 accepts an input image, then downscales the image's feature maps while obtaining high-level semantic features in the process.

Consequently, to restore the feature map's resolution, the decoding structure combines fine-scale details obtained from the lower layers of the encoding structure with the high-level features through extensive residual connections to generate effective feature expression. Moreover, the combined encoder-decoder structure captures long-range semantics critical for segmentation accuracy. In Encoder 1, five structural blocks gradually reduce the image dimensions as the image channel features increase, raising the high-level image semantics.

The SDRNet employs the ResNet50 as the backbone of our network due to its remarkable feature extraction capabilities. We use pre-trained ResNet50 on Encoder 1 since pre-trained networks significantly enrich the network's basic feature learning capabilities and reduce the CNNs demand for massive training labels. However, since the RS scenes are different from natural images in the originally trained dataset, we disable the pretrained setting in the second encoder to make the network learn more indicative features. This helps to activate deeper layers to learn more high-level features effectively. The symmetrical encoders and decoders stacked together work as spatial reconstruction subnets where rich spatial details are encoded, and low-level feature maps (from lower layers) are generated. Notably, both Encoder and Decoders in the first subnetwork are symmetrical and use five structural blocks containing deconvolution and up-sampling layers.

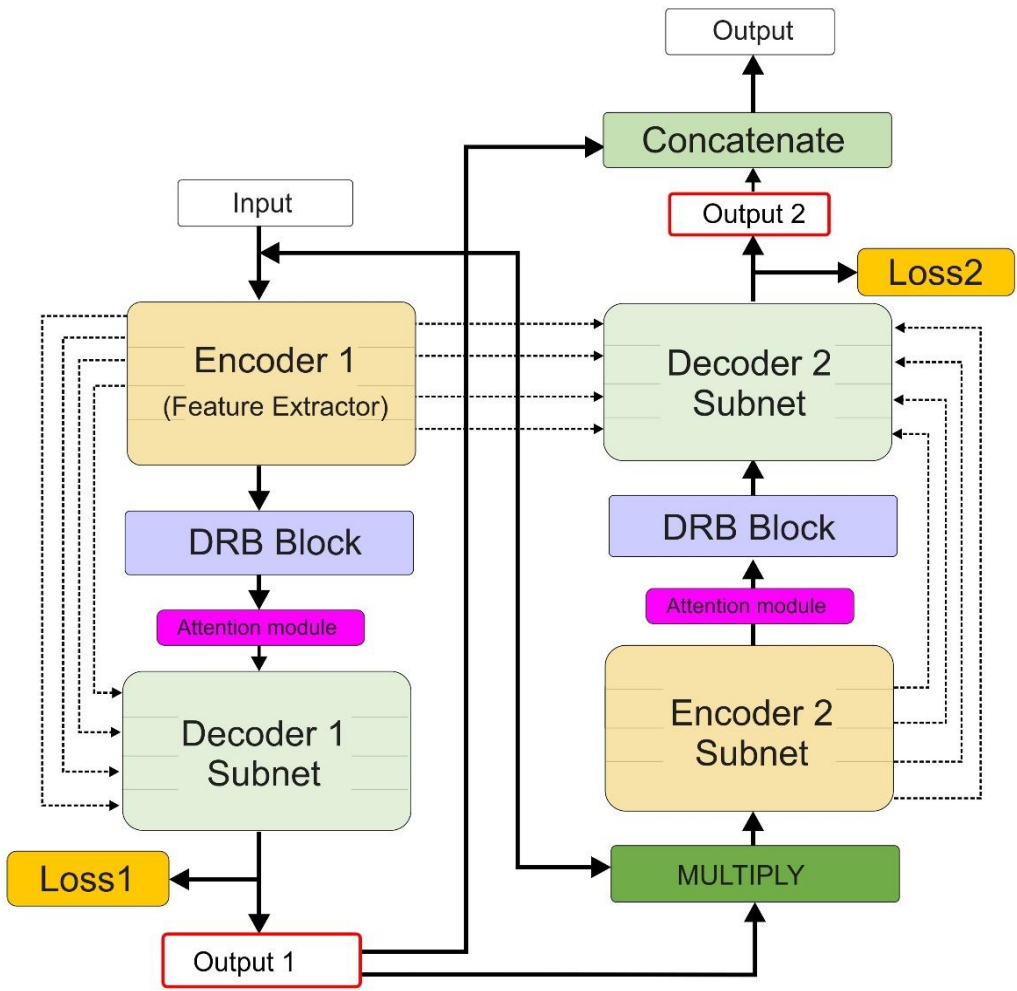

**Fig. 2**. An illustration of the proposed SDRNet framework

The second subnetwork consists of a similar design with some modifications. Unlike the first sub-network, the second Encoder and Decoder are modified to have only four structural blocks to minimize the number of learnable features and reduce the network complexity. The two subnetworks are connected in a hierarchical fashion where the output of Decoder 2 is input to Encoder 2 for deeper feature learning and extraction.

Before connecting to a multi-class softmax function, both decoders unsampled the size to the initial image dimensions.

In addition, since RS data contains redundant and undesired features not included in the output classes (referred to as stuff), the attention mechanism handles the amorphous and redundant features. Inspired by the human visual system, the self-attention mechanism refines the network to focus on specific regions of importance, thus significantly minimizing the cost of learning features from the undesired areas and redundant data. Besides, since it is practically impossible to activate and learnable parameters in DCNNs, the network constrains features and only harnesses class-specific features.

Attention mechanisms have been extensively used in semantic segmentation tasks (Li et al., 2019; Li et al., 2021b; Zhong et al., 2020) and have shown significant performance improvements by refining the learnable features from images. We construct self-attention blocks with input from layers 3, 4, and 5 of Encoder 1 and layers 2, 3, and 4 of Encoder 2 to extract multi-level features.

The self-attention mechanism used in our network is defined in:

$$M_s(F) = \sigma(F^{axa}([\text{AvrPool}(F); \text{MaxPool}(F)])), \quad (1)$$

where σ denotes the sigmoid function, and $F^{a \times a}$ represents a convolution operation with the filter size of a × a.

In addition, our framework employs a residual learning strategy to overcome the vanishing gradient problem. He et al. (2016) revolutionary concept of residual learning is now a mature concept that yields superior performance in various image analysis tasks, especially in constructing deeper networks (Zhong et al., 2017). The identity mapping replaces the stacked convolutional blocks, which helps build deeper networks unaffected by the vanishing gradient problem. Following skip connections, network gradients can flow through directly without passing through non-linear activation functions, thus alleviating gradient explosion or vanishing. Moreover, skip connections improves gradient flow during backpropagation, accelerating the convergence in deeper networks. Following this intuition, information can flow from the encoding phase of our network to the matching pairs in the decoding phase. The progressive concatenation of fine-location-based details improves the reconstruction of the original image dimensions. The residual function is defined in:

$$y = F(x, Wi) + x, \quad (2)$$

where *x* represents the input, y represents the output feature maps, and *F(x, Wi)* function represents the residual connection. The results *F(x, Wi)* should have the exact dimensions as *x*.

Moreover, we adopt the intuition of residual connections to facilitate information flow in the dilated residual module. Residual connections support the building of deeper networks where low, mid, and high-level features can be harnessed and fused to generate the necessary robust feature descriptor.

*3.3 Multicontextual feature aggregation using DRB blocks*

Multicontextual information is critical for accurate segmentation. Primarily, most image classification models integrate multi-scale contextual information via successive downsampling through pooling operations and sub-sampling layers. Such actions reduce the resolution to attain global prediction (Krizhevsky et al., 2017; Simonyan and Zisserman, 2014). On the other hand, pixel-level classification assumes the reasoning of the multi-scale context with the matching of the input-output resolutions. The step size of the kernel is defined by the stride parameter in CNNs when traversing the image. The default stride value is 1; however, a value less than 1 can be used for down-sampling an image, similar to a max-pooling operation. Convolutional layers with a stride greater than 1 reduce the spatial resolution making pixel-level prediction less accurate. The alternative to obtaining large RF without using striding layers is by using larger kernels. The main challenge of such an approach is the computational complexity and increased parameters, which makes most networks slower and more susceptible to overfitting.

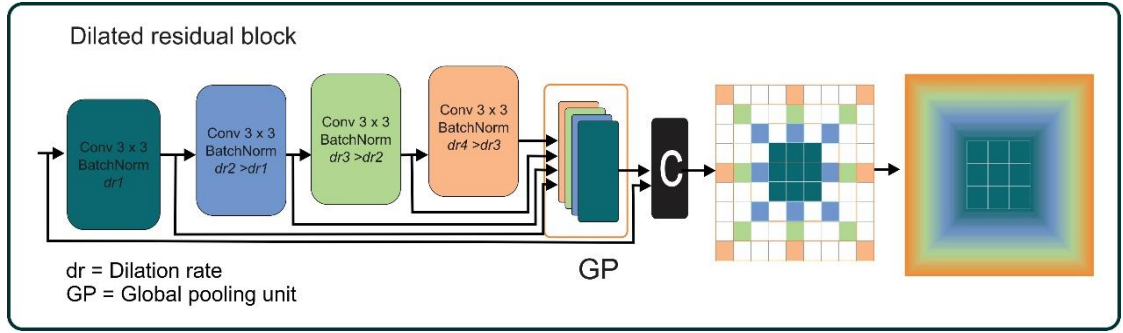

**Fig. 3**. Dilated residual blocks

To obtain multi-contextual features, dilated convolution (Yu and Koltun, 2016) introduced large sparse kernels in pooling and convolution layers to attain multi-scale contextual information aggregation without sacrificing the image resolution. Dilated convolutions are considered the better option and have successfully demonstrated improved performance in many classification and segmentation tasks (Li et al., 2018; Wei et al., 2018). A 2D dilated convolution operator is defined in:

$$y(m, n) = \sum_{i=1}^{M} \sum_{j=1}^{N} (m + r \times i, n + r \times j) w(i, j), \qquad (3)$$

where *y(m, n), x(m, n),* and *w(i, j)* are the input, output, and the filter size of *M* and *N* dimensions of the dilated convolution, respectively, the parameter *r* represents the dilation rate. The dilated convolution acts normally when the dilation rate is equal to

1. A dilation rate less than 1 denotes detailed segmentation of the feature map, which takes more training time. At the same time, larger than 1 causes an enlargement in the RF without raising the number of parameters or computation requirements. Different dilation rates can be used to adjust the RF range.

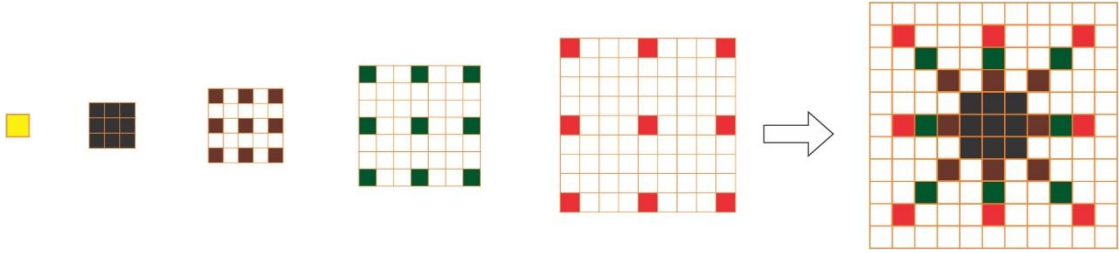

**Fig. 4.** Illustration of the resultant kernel of the dilated residual block (DRB).

Inspired by dilated convolution concept, we proposed a dilated residual block that uses progressive dilated rates in different layers to achieve full RF without loss of resolution or coverage. The goal of the DRB block illustrated in **Fig. 3** is to raise the RF to capture global dependencies sufficiently. However, large dilation rates can cause a gridding effect that significantly inhibits the dilated layers' performance. The gridding problem results from not all pixels getting kernel responses resorting to a "checkerboard effect" and loss of continuity of information (Hou et al., 2021). This is harmful to the task of pixel-level dense prediction.

To solve this, DRB uses dilated layers with progressive dilation rates. The layers are extensively connected progressively via skip connections to enhance information flow between the dilated layers. This significantly overcomes the gridding issue by ensuring that the resultant dilated kernel obtains responses from all the regions in an image without increasing the number of kernel parameters. The resulting dilated kernel is shown in **Fig. 4.** The goal of the DRB is to cover all holes and can be defined in:

$$M_i = \max[M_{i+1} - 2r_i, M_{i+1} - 2(M_{i+1} - r_i), r_i, \qquad (4)$$

where $r_i$ is the dilation rate of the i layer and $M_i$ refers to the maximum generation rate at the *i* layer, and n is the total number of layers.

### *3.4 Training supervision using intermediate loss*

Classical DCNNs perform the final prediction at the end of the entire network. The backpropagation process learns from the later and deeper network layers to the former. However, in very deep networks, the vanishing gradient problem can impede the effective learning of all the layers from deep to shallow ones. In the case of SDRNet, stacking two networks can cause irredeemable gradient deterioration. An intermediate loss can improve the gradient flow and enhance the learning during the backpropagation process to overcome gradient decay. Moreover, this method can effectively guide the middle layers by computing how bad the model performs compared to the ground truth labels.

The SDRNet model deploys two losses where each loss is connected to the ground truth labels. Introducing two loss functions at the end of each network ensures the outputs are consistent with the ground truth and further guides the backpropagation midway through the network. Loss 1 is computed to infer how wrong the prediction is from the desired ground truth multi-class label, thus optimizing the learning process and acting as intermediate supervision. Loss 1 is defined in :

$$Loss\ 1 = -\frac{1}{N}\sum_{i=1}^{N} W_i . p_i . log\left(\frac{e_{pi}}{\sum_{j=1}^{N} e^{pj}}\right), \tag{5}$$

where $N$ is the number of classes, $W_i$ represents the class weight $i$, and $p_i$, $pj$ denotes the predictions and the ground truth distribution of class $i$, respectively. We define the total loss in:

$$L_T = (\alpha \times MainLoss) + (\beta \times InterL1), \tag{6}$$

where α and β are the respective weights in our network, and Loss1 and MainLoss are loss values of the output layer and the end of the first network, respectively.

## 4. Experiments and analysis

### 4.1 Datasets descriptions

We use the benchmark datasets of ISPRS Potsdam and Vaihingen land cover datasets to test our framework's performance and effectiveness. Two reasons informed the choice of the datasets: Being benchmark datasets and having required high-resolution images. More details on the datasets are available on the dataset website (http://www2.isprs.org/commissions/comm3/wg4/semantic-labeling.html).

**Vaihingen dataset:** The Vaihingen semantic labeling dataset contains 33 images of 2494 × 2064 pixels and a ground truth resolution of 5 cm. The dataset also has red, green, and NIR channels provided together with DSM. Of the 33 images, 16 and 17 images are set for the training and test set. In our experiments, we used image with IDs: '1', '3', '5', '7', '11', '13', '15', '17', '21', '23', '26', '28', '30', '32', '34', '37' for training, images with IDs '7', '28' for validation, and images with IDs '2', '4', '6', '8', '10', '12', '14', '16', '20', '22', '24', '27', '29', '31','33', '35', '38' for testing which are illustrated in **Fig.5**. The DSM channels were not considered in our experiments.

**Potsdam dataset**: The Potsdam landcover dataset has 38 VHR images of 6000 × 6000 pixels with a ground truth resolution of 5 cm. Provided in the dataset are NIR, red, green, and blue channels,  and also the DSM and normalized DSM (NDSM). We set the dataset as follows: Training set with image IDs: : '2_10', '2_11', '2_12', '3_10',

'3_11', '3_12', '4_10', '4_11', '4_12', '5_10', '5_11', '5_12', '6_10', '6_11', '6_12', '7_10', '7_12', '7_11', '6_7', '6_8', '6_9', '7_7', '7_8', and '7_9'. Testing set with images IDs: testing, ID: '2_13', '2_14', '3_13', '3_14', '4_13', '4_14', '4_15', '5_13', '5_14', '5_15', '6_13', '6_14', '6_15', '7_13' , and validation set with image IDs: '4_10', '7_10' as shown in **Fig. 5.** We used only used the ground truth images with eroded boundaries as provided by the ISPRS. Our experiments did not use nDSM channels. All the images and corresponding masks were cropped to size 256 x 256, and augmented using the albumentations library (Buslaev et al., 2018a). The package supports transformations such as flipping, rotation, scaling, transposition, shift scaling, grid distortion among others.

From the original color map, the six land cover classes are defined as follows: Buildings (blue), Trees (green), Low vegetation (cyan), Clutter (red), Impervious surfaces (white), Cars (yellow), and Undefined (black). The clutter class includes some water bodies and other incoherent objects.

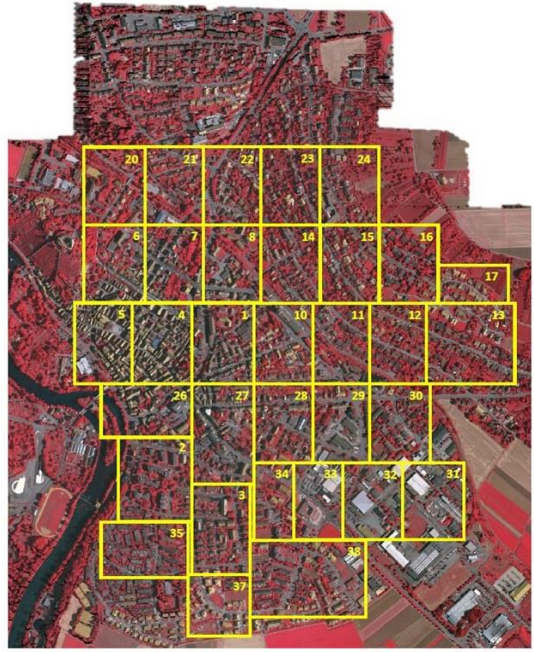

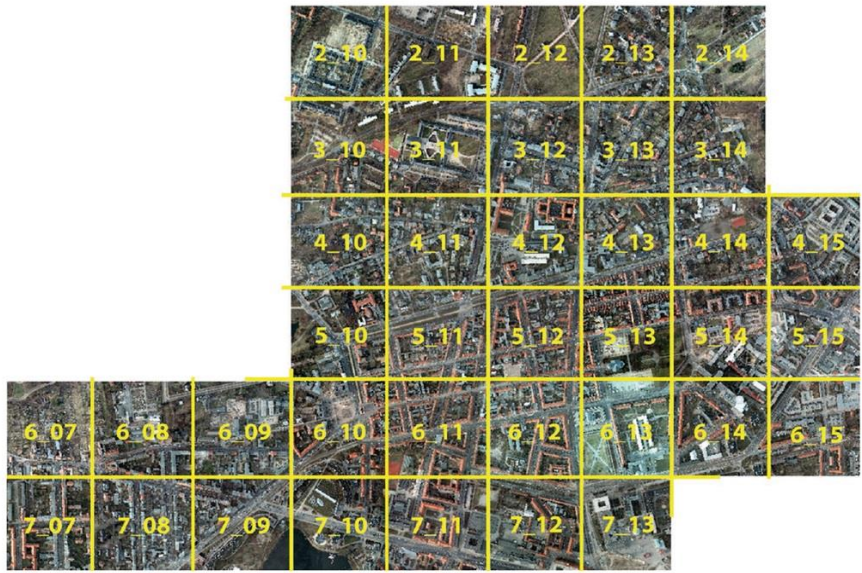

**Fig. 5.** Outlines of the image patches given in the Vaihingen region and Potsdam city for 2D semantic labeling overlaid with the true orthophoto mosaic. Numbers refer to the individual patch numbers

*4.2 Training details and configuration*

Our experiments used the Pytorch framework on the NVIDIA GeForce RTX 2080Ti GPU graphics card, 11 GB RAM, and CUDA 10.2 and Cudnn7.6.5. The platform was used for training and testing our framework. We used Adam optimizer (Kingma and Ba, 2014)with AMSGrad (Reddi et al., 2018) and weight decay of $2\times 10^{-5}$. The polynomial decay rate was set at $(1 - (cur\ iter/max\ iter))\ 0.9$. We used an input image of $256 \times 256$ pixels and a batch size of 5 due to the limitation of the computational power.

Image pre-processing can improve training speed and enhance the network's self-fitting process, especially when the two datasets have insufficient training samples to train data-hungry supervised DCNN models. Data augmentation was used to supplement the training set in our experiments using the albumentations library (Buslaev et al., 2018a). The package provides several flexible and efficient image augmentation functions relating to color, contrast, brightness, and other geometric transformations. The package supports transformations such as flipping, rotation, scaling, transposition, shift scaling, and grid distortion.

*4.3 Evaluation metrics*

The following assessment metrics were used to measure the performance of our proposed framework. OA, mF1, precision, recall, and specificity, which are computed as:

$$OA = \frac{\sum_{k=1}^{K} TP_k}{\sum_{k=1}^{K} TP_k + FP_k + TN_k + FN_k}, \tag{7}$$

$$Precision = \frac{TP_k}{TP_k + FP_k}, \tag{8}$$

$$Recall = \frac{TP_k}{TP_k + FN_k}, \tag{9}$$

$$Specificity = \frac{TN_k}{TN_k + FP_k}, \tag{10}$$

$$F1\ Score = 2 \times \frac{Precision_k \times Recall_k}{Precision_k + Recall_k}, \tag{11}$$

where $TP_k$, $FP_k$, $TN_k$, and $FN_k$ represent the true positive, false positive, true negative, and false negatives, respectively, for an image of class k. *OA* and *mF1* represent overall accuracy and *mean F1* (The average F1 score for all classes). Specificity evaluates the model's ability to predict $TN_k$ for an image of class k. We used all classes but clutter to compute the OA. The clutter class was ignored since it has minor representation in the training set.

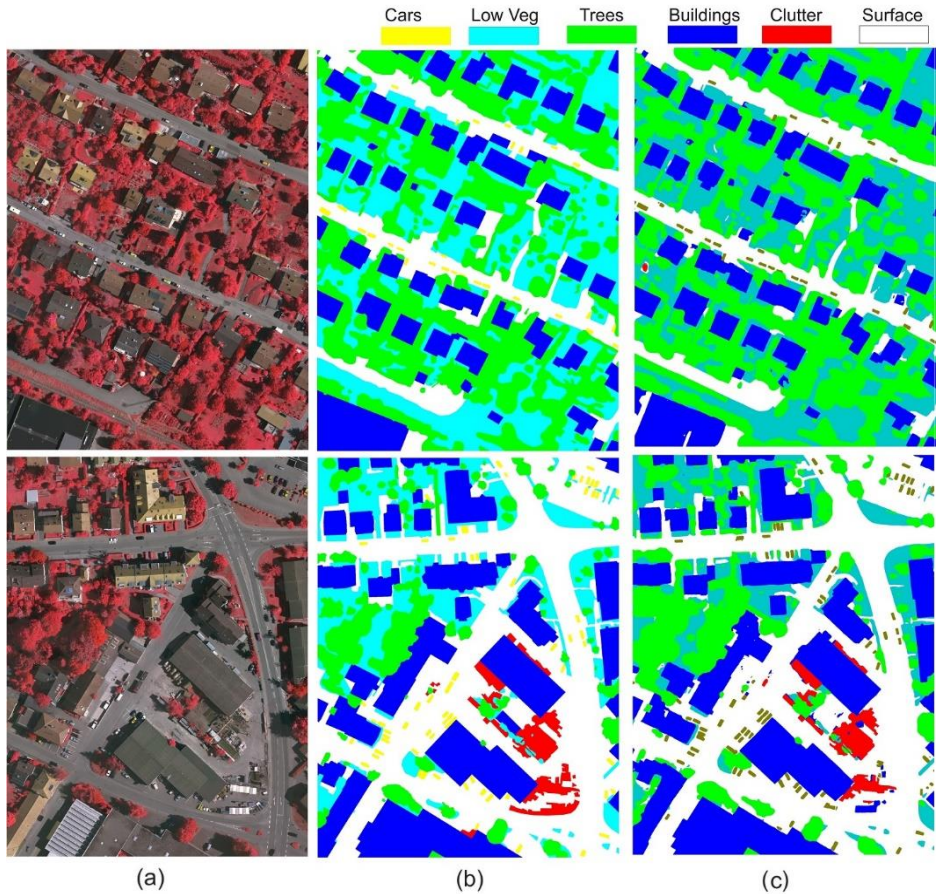

**Fig. 6.** Segmentation results for the Vaihingen dataset of image tiles 12 and 29. (a) = input image tile, (b) = ground truth, and (c) = prediction.

## 4.3. Results and discussions

This section demonstrates our proposed SDRNet's capability to preserve the essential spatial details using the cascaded encoder-decoder scheme and harnessing multi-contextual information from different levels of the encoder backbone using DRB modules. Later, the role of individual components used in our framework is analyzed in the ablation studies.

**Table 1**

**Per class results (in percentage) on the test datasets of ISPRS Potsdam and Vaihingen trained with the SDRNet model**

| Per class classification scores on the Potsdam dataset | | | | | | |
|---|---|---|---|---|---|---|
| | *Buildings* | *Trees* | *Low veg.* | *Road surf* | *Cars* | *Mean F1* |
| ***Specificity*** | 97.75 | 97.60 | 96.61 | 96.38 | 99.86 | 97.64 |
| ***Precision*** | 93.50 | 90.75 | 87.84 | 94.78 | 92.65 | 91.90 |
| ***Recall*** | 97.64 | 90.25 | 89.25 | 93.56 | 97.64 | 93.67 |
| ***F1 score*** | 95.53 | 90.50 | 88.54 | 94.17 | 95.08 | 92.76 |

*Per class classification scores in the Vaihingen dataset*

|  | *Buildings* | *Trees* | *Low veg.* | *Road surf* | *Cars* | *Mean F1* |
|---|---|---|---|---|---|---|
| ***Specificity*** | 97.75 | 97.60 | 96.61 | 96.38 | 99.86 | 97.64 |
| ***Precision*** | 93.50 | 88.58 | 87.16 | 92.13 | 91.71 | 90.62 |
| ***Recall*** | 97.64 | 90.33 | 88.50 | 92.08 | 97.85 | 93.28 |
| ***F1 score*** | 95.53 | 89.45 | 87.82 | 92.10 | 94.68 | 91.93 |

The per-class results of the SDRNet model are shown in **Table 1.** The SDRNet model achieves an F1 score of between 88% and 95% in all classes on the Potsdam dataset and between 87% and 95% in the Vaihingen land cover dataset. The two labeling benchmark datasets are complex and contain highly heterogeneous class objects of varying sizes. Moreover, some land cover class objects have the shadow effect, significantly inhibiting accurate segmentation.

Qualitative classification results for Vaihingen and Potsdam datasets are presented in **Figs. 6** and **7**. Despite the complexities mentioned above in the two datasets, our proposed model can correctly classify most land cover objects accurately. For example, the car class attained a competitive F1 score of about 95% in both datasets. This class is considered challenging to classify accurately due to its size limitation. The accurate scores from small-sized objects can be attributed to the successive spatial information restoration using multi-level low-level features obtained from the encoding phases of the dual encoder-decoder scheme. Specifically, the reconstruction of spatial details supports the accurate classification and delineation of small-sized objects. Moreover, from the segmentation maps, smaller class objects such as trees have consistent pixels and are clearly labeled, which signifies our model's reconstruction power of spatial details for correct segmentation.

In addition, our proposed model attains an F1 score of approximately 96% in the building class, despite this class possessing shadow effect, which can affect the segmentation results.

**Table 2**
**Comparison of OA and mF1 performance between SDRNet model and other published methods on the ISPRS Potsdam landcover dataset test set. (Scores indicated in %)**

| **Models** | **OA** | *Imp. Surf* | *Buildings* | *L.Veg* | *Trees* | *Cars* | **mF1** |
|---|---|---|---|---|---|---|---|
| DANet (Fu et al., 2019) | 89.1 | 91.0 | 95.6 | 86.1 | 87.6 | 84.3 | 88.9 |
| BiSeNetV2 (Yu et al., 2018) | 88.2 | 91.3 | 94.3 | 85.0 | 85.2 | 94.1 | 90.0 |
| ShelfNet (Zhuang et al., | 89.9 | 92.5 | 95.8 | 86.6 | 87.1 | 94.6 | 91.3 |
| 2019) | | | | | | | |
| HCA Net Res101 (Bai et al., 2022) | 88.9 | 91.4 | 94.8 | 87.1 | 84.1 | 91.9 | 88.1 |
| CDR-Net (Wambugu et al., 2021) | 90.7 | 92.9 | **96.7** | 87.4 | 88.6 | 94.8 | 92.1 |
| **SDRNet** | **90.8** | **94.2** | 95.5 | **88.5** | **90.5** | 95.1 | **92.8** |

To demonstrate the efficiency of our framework, we compared our models with other models based on the encoder-decoder paradigm, multi-contextual information aggregation, attention mechanisms, and other current approaches. The comparative experimental results are presented in **Tables 2 and 3** for Potsdam and Vaihingen datasets**.** The scores marked in bold represents the best per class and overall accuracy from the compared models.

The results show that our model posted higher scores in the mean F1 and OA significantly compared to the other models in most classes on both land cover datasets. Specifically, our model can classify the "car," "low vegetation," and "buildings" classes better in the Vaihingen dataset compared to the other models. The car class is challenging since its objects are relatively small, and it is considerably challenging to attain improvements on existing established models, especially on ISPRS labeling benchmarks.

We compared our model against ShelfNet (Zhuang et al., 2019), which follows a similar intuition of multiple encoder-decoder branch pairs with skip connections at each spatial level. The SDRNet performs better in all but the building class in the Potsdam dataset based on the numerical scores. The improvement in our model can be explained by the multi-contextual information provided by the DRB block in our network structure. Still, compared to the dual-attention mechanism-based model DANet (Fu et al., 2019), we note that although it is ResNet backbone with two last layers dilated, our model achieves an OA improvement of 1.7% and 2.4% on the Potsdam and Vaihingen datasets.

**Table 3**
**Comparison of OA and mF1 performance between SDRNet model and other published methods on the ISPRS Vaihingen landcover dataset test set. (Scores indicated in %)**

| *Models* | *OA* | *Imp. Surf.* | *Buildings* | *L.Veg* | *Trees* | *Cars* | *mF1* |
|---|---|---|---|---|---|---|---|
| RefineNet (Lin et al., 2017) | 84.4 | 87.6 | 88.5 | 81.9 | 79.1 | 87.9 | 85.0 |
| DANet (Fu et al., 2019) | 88.2 | 90.0 | 93.9 | 82.2 | 87.3 | 44.5 | 79.6 |
| MAResU-Net (Li et al., 2022a) | 90.1 | 92.0 | 95.0 | 83.7 | 89.3 | 78.3 | 87.7 |
| HCA Net Res101 (Bai et al., 2022) | 89.7 | 92.2 | **95.6** | 80.7 | 88.9 | 87.4 | 88.9 |
| CDR-Net (Wambugu et al., 2021) | 90.5 | **92.7** | 95.4 | 83.4 | 89.6 | 88.7 | 90.0 |
| *SDRNet* | **90.6** | 92.1 | 95.5 | **87.8** | 89.5 | **94.7** | **91.9** |

The contextual information harnessed by the DRB in our proposed model significantly supports the correct classification of both large and small-sized objects. Semantic information captured at different levels of our network's backbone and later fused into the DRB is essential for accurate scores. The attention mechanism achieved through spatial blocks significantly improves feature learning, weighing more on crucial regions and less on insignificant ones.

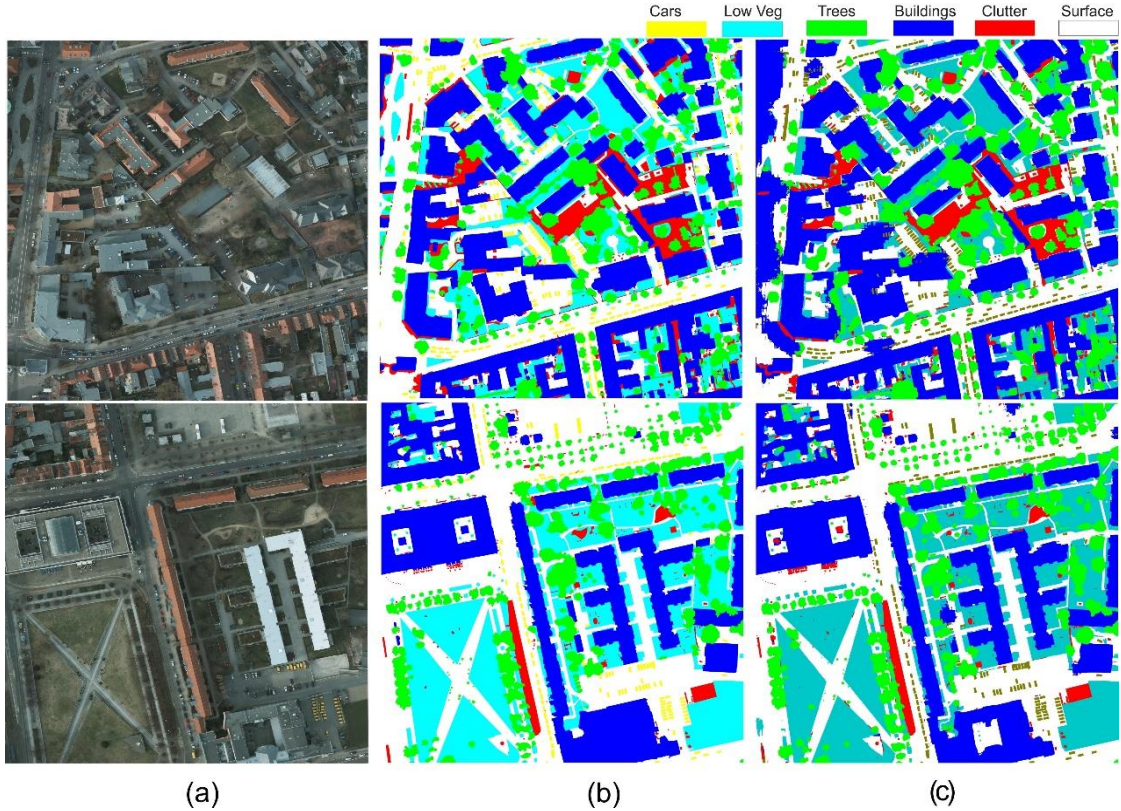

**Fig. 7.** Segmentation results of the Potsdam dataset image tiles 4-13 and 6-13. (a) = input image tile, (b) = ground truth, and (c) = predictions.

However, regions with shadows pose some significant challenges to accurate classification in our model. For instance, the shadows cast in the buildings limit our model's accuracy scores in both datasets scoring second last on the Potsdam dataset. A possible solution can be developing a branch to handle the shadow features such that the areas of interest are delineated and learned without the shadow regions. This remains a possible research focus in our future experiments and study. The ablation studies section discusses the significance of each part of the SDRNet model in detail.

*4.5 Ablation studies*

**Comparison with former CRD-Net framework.**

The previous work proposed a CRD-Net framework (Wambugu et al., 2021) for land cover classification using FRRS images in the previous work. The framework follows a similar intuition of the encoder-decoder with some improvements to the network design. For instance, the SDRNet uses two encoders-decoder subnetworks stacked together to form a lightweight network with a single attention block on each encoder compared to a pair in the CRD-Net model.

In addition, the SDRNet model uses five and four residual blocks on the first and the second encoders, respectively, significantly minimizing the number of learnable parameters and reducing network complexity. The same design is affected in the first and the second decoders in the network. Moreover, the intermediary loss in the SDRNet model is connected between two subnetworks, while in the former CRD-Net model, the intermediate loss is connected between the 'network's encoding and decoding phases. In both models, the main loss is connected at the end of the network. In addition, the CRD module is different from the DRB module used in the SDRNet. The latter connects dilated layers hierarchically and concatenates all the resultant feature maps at the end using a feature fusion block. In contrast, the CRD module connects the input from the dual attention blocks to each dilated layer in a parallel format. In both cases, dense concatenation is done at the end of the dilated layers.

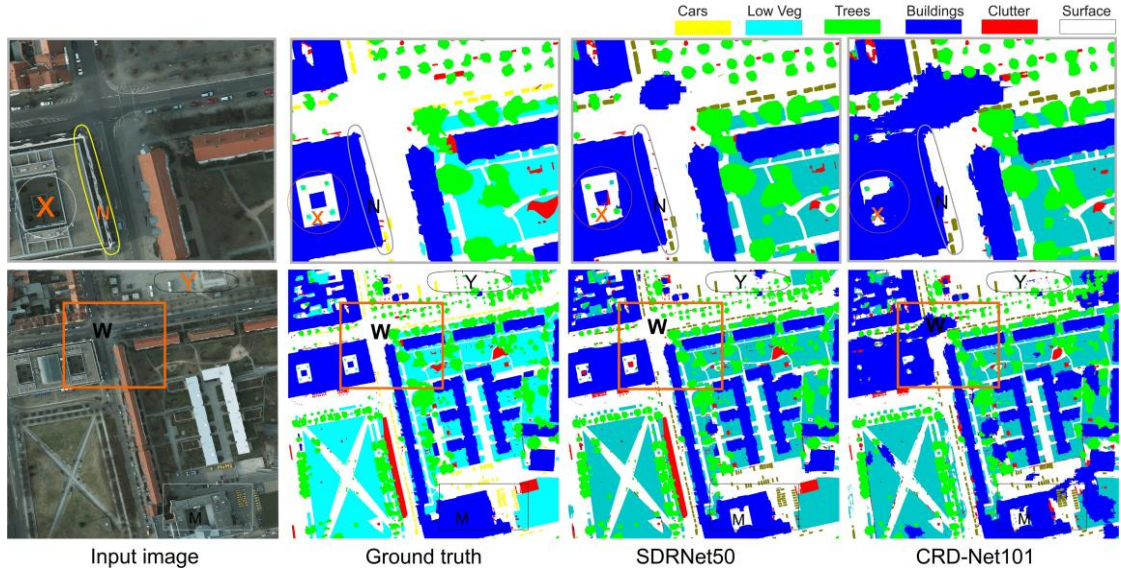

**Fig. 8.** Comparison results between CRD-Net and SDRNet on the Potsdam dataset.

The last two rows of **Tables 2 and 3** show the quantitative performance comparison between the SDRNet and the CRD-Net models. Based on the per-class experimental results, the SDRNet model performs better than the CRD-Net model in the road surface, low vegetation, trees, and the car class in the Potsdam dataset. Still, on the Vaihingen dataset, the SDRNet outperforms the CRD-Net in all the classes but the trees class, where the two models achieve the same F1 score of 89.6%. The improved performance can be attributed to high-level semantic features extracted in the stacked network and integrating the loss after reconstruction of spatial details at the end of the first subnetwork, rather than between the encoder and the decoder as in the case of the CRD-Net model.

The SDRNet improves the CRD-Net by an OA of 0.11% on the Vaihingen dataset, although CRD-Net uses a more computationally demanding ResNet101 backbone with five residual blocks on the encoder. The ResNet101 backbone used in CRD-Net has more parameters (123.14MB) than the ResNet50 used in the SDRNet (85.15MB). The model's complexity and parameters comparison of the two models is shown in **Table 4.** Still, the SRDNet model is faster and uses only about 70% of the FLOPS and 60% of the inference time used in the CRD-Net. Overall, the SDRNet framework posted better accuracy scores and is computationally efficient than the CRD-Net model.

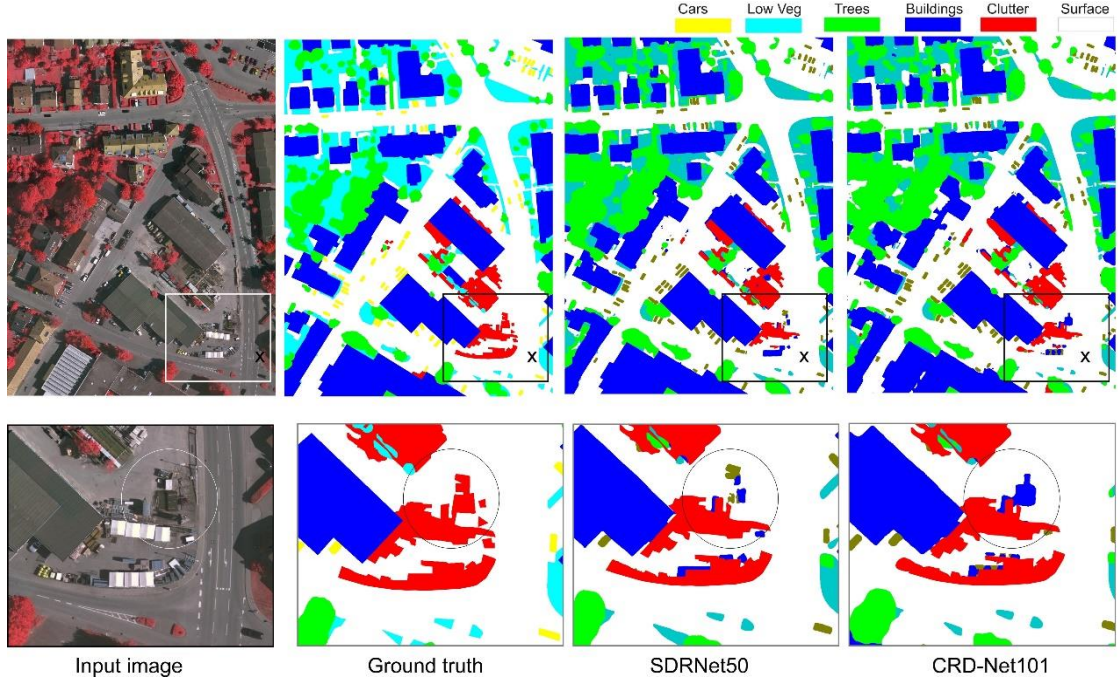

**Fig. 9.** Comparison results between CRD-Net and SDRNet on the Vaihingen dataset image.

We demonstrate the qualitative performance comparison between the CRD-Net and the SDRNet on image tile 6-13 of the Potsdam dataset in **Fig. 8**. As visible in the figure, the region marked *W* is enlarged on the top to demonstrate the visual performance of each model qualitatively. Specifically, results from the SDRNet model show better edge delineation in large objects such as buildings (see areas on the enlarged region marked *X* and *W*) and a more accurate delineation of surfaces than CRD-Net. (Refer to regions marked as *N* and *M*).

In addition, **Fig. 9** shows the visual results of the comparison between the SDRNet and the CRD-Net models tested on the image tile 29 of the Vaihingen datasets. The region marked X at the top image is enlarged in the image at the bottom of **Fig 9**. Visibly, the encircled region has highly homogeneous objects, with clutter being the only identifiable class in the ground truth. However, other "stuff" objects in the region have visual appearances and textual features closely related to other classes, such as buildings and cars, making the two-class objects wrongly classified in the identified region. Such instances can be challenging to even complex frameworks.

**Table 4**
**CRD-Net and SDRNet model's complexity and parameters tested on the Vaihingen dataset**

| Model base network | Parameters | FLOPS | Speed | OA |
|---|---|---|---|---|
| SDRNet50 | 38.64 MB | 18.72GB | 14.21 ms/iter | 86.2 |
| CRD-Net101 | 123.14MB | 73.39GB | 19.57ms/iter | 84.9 |
| SDRNet18 | 28.20MB | 5.77GB | 11.12ms/iter | 85.2 |
| CRD-Net50 | 85.15 MB | 53.94GB | 16.71 ms/iter | 85.8 |

However, the task of full automation of land cover classification using FRRS images can only be considered as some work in progress. The proposed frameworks based on the hybrid model and stacked network still have chances for further improvements to realize higher and more competitive levels. For instance, there are several regions where different class objects are highly homogeneous, making the separation and extraction of features difficult (See region marked Y). Moreover, regions with shadow cast and where some objects cause occlusion to neighbouring objects make accurate classification more difficult. These are areas subject to further research.

**Ablation experiments**

To better evaluate the effectiveness of the components in the proposed SDRNet, we conducted ablation experiments on the Vaihingen dataset under different settings. The training setup for all various configurations was the same. The details of the settings and the results are shown in **Table 5.**

**Table 5**
**Performance evaluation of the SDRNet model under different configurations on the Vaihingen landcover dataset. (Scores indicated in %)**

| Configuration | OA | Buildings | Trees | L.Veg | Imp. Surf. | Cars | mF1 |
|---|---|---|---|---|---|---|---|
| No attention blocks | 89.6 | 94.2 | 88.9 | 86.2 | 91.6 | 90.5 | 90.1 |
| No DRB | 87.8 | 92.1 | 87.7 | 84.9 | 90.5 | 90.2 | 89.1 |
| Single Network | 87.9 | 93.5 | 88.9 | 85.2 | 90.3 | 91.6 | 89.9 |
| SDRNet18 | 90.2 | 93.8 | 88.6 | 85.8 | 91.5 | 93.2 | 90.6 |
| SDRNet50 | 90.6 | 95.5 | 89.5 | 87.8 | 92.1 | 94.7 | 91.9 |

**SDRNet50:** The proposed framework comprises two stacked subnetworks, each with an encoder and decoder and a DRB between the downsampling and upsampling paths using an improved ResNet50 backbone in both subnetwork.
 **No attention blocks**: The SDRNet was implemented without attention blocks in both encoders.
**No DRB**: The SDRNet was implemented without the DRB in both stacked subnetworks.
**SDRNet18**: SDRNet implemented on a ResNet 18 backbone.
**Single network**: SDRNet is implemented on a single encoder and single decoder.

The results demonstrate the significance of various components forming the stacked network. Specifically, training the model without the DRB results in a significant OA loss of 2.8%. In addition, Smaller objects such as the "car" class are critically affected by the absence of the multi-contextual information harnessed by the DRB component. When the network was tested without the DRB block, the F1 scores of the car and low vegetation class reduced by 4.5% and 2.9%, respectively.

Also, the influence on multi-level feature extraction using stacked networks can be demonstrated in the single network results. Under this setting, only a single encoder and a decoder was used (Subnetwork 1). The results can demonstrate the influence of an elaborate multi-level feature learning scheme in harnessing multi-level image features. Using only a single network significantly reduces the OA by 2.7%.

Later, the influence of spatial attention blocks is investigated. The attention blocks help the model discriminate features in some image areas while assuming less critical regions. Elimination of the attention block reduced the F1 score by 4.5% and 2.3% in car and buildings classes, demonstrating the influence of the spatial attention mechanism on the network performance. The results under the ResNet18 backbone indicate that our model is robust and can still perform significantly well with the light network and compares favorably well to other lightweight models under investigation. More ablations on the impact on intermediary loss, different loss functions, and different backbone networks are anticipated in future experiments.

## 5. Conclusion

In this study, we proposed a novel lightweight SDRNet model to handle the delicate task of semantic segmentation of FRRS data. Our framework uses two stacked subnetworks, each comprising an encoder and a decoder. The encoder paths harness sufficient spatial details from the fine-resolution RS images, and the decoder path helps recapture the spatial information through extensive skip connections. The DRB captures the necessary multi-contextual information between encoder and decoder by enlarging the RF through successive dilated layers. In addition, the SDRNet framework uses an intermediate loss function between the first and the second subnetworks to guide the training process at the middle layers of the network. Subsequent residual connections implemented in the DRB significantly avoid the gridding effect prone to dilated convolutions. Extensive experiments show that our framework competes favorably against other compared models for the semantic segmentation task using FRRS images. Future work will incorporate multimodal data (LiDAR and DSM) and test our framework using other backbone networks.

**Declaration of Competing Interest**

The authors declare no compelling interests that could have influenced the research reported in this paper.

**Funding**

This work was supported by the National Key Research and Development Program of China（2022YFB2602105）and the Key Technological Innovation Program of Ningbo City under Grant No. 2024Z297.

Rezaee, M., Mahdianpari, M., Zhang, Y., Salehi, B., 2018. Deep Convolutional Neural Network for Complex Wetland Classification Using Optical Remote Sensing Imagery. IEEE J. Sel. Top. Appl. Earth Obs. Remote Sens. 11, 3030-3039.

Ronneberger, O., Fischer, P., Brox, T., 2015. U-Net: Convolutional Networks for Biomedical Image Segmentation. 2015 Conference on Medical Image Computing and Computer-Assisted Intervention. MICCAI, 9351, 234-241.

Samie, A., Abbas, A., Azeem, M.M., Hamid, S., Iqbal, M.A., Hasan, S.S., Deng, X., 2020. Examining the impacts of future land use/land cover changes on climate in Punjab province, Pakistan: implications for environmental sustainability and economic growth. Environmental Science and Pollution Research 27, 25415-25433.

Samy, M., Amer, K., Eissa, K., Shaker, M., Elhelw, M., 2018. NU-Net: Deep Residual Wide Field of View Convolutional Neural Network for Semantic Segmentation. 2018 IEEE/CVF Conference on Computer Vision and Pattern Recognition, pp. 267-271.

Simonyan, K., Zisserman, A., 2014. Very Deep Convolutional Networks for Large-Scale Image Recognition.

Volpi, M., Tuia, D., 2017. Dense Semantic Labeling of Subdecimeter Resolution Images With Convolutional Neural Networks. IEEE Trans. Geosci. Remote Sens. 55, 881-893.

Wambugu, N., Chen, Y., Xiao, Z., Wei, M., Aminu Bello, S., Marcato Junior, J., Li, J., 2021. A hybrid deep convolutional neural network for accurate land cover classification. Int. J. Appl. Earth Obs. Geoinf. 103, 102515.

Wang, F., Jiang, M., Qian, C., Yang, S., Li, C., Zhang, H., Wang, X., Tang, X., 2017. Residual Attention Network for Image Classification, 2017 IEEE Conference on Computer Vision and Pattern Recognition (CVPR), pp. 6450-6458.

Wei, Y., Xiao, H., Shi, H., Jie, Z., Feng, J., Huang, T.S., 2018. Revisiting Dilated Convolution: A Simple Approach for Weakly- and Semi-Supervised Semantic Segmentation, 2018 IEEE/CVF Conference on Computer Vision and Pattern Recognition, pp. 7268-7277.

Wu, Z., Shen, C., van den Hengel, A., 2019. Wider or Deeper: Revisiting the ResNet Model for Visual Recognition. Pattern Recognition 90, 119-133.

Yang, X., Li, S., Chen, Z., Chanussot, J., Jia, X., Zhang, B., Li, B., Chen, P., 2021. An attention-fused network for semantic segmentation of very-high-resolution remote sensing imagery. ISPRS J. Photogramm. Remote Sens. 177, 238-262.

Yi, Y., Zhang, Z., Zhang, W., Zhang, C., Li, W., Zhao, T., 2019. Semantic Segmentation of Urban Buildings from VHR Remote Sensing Imagery Using a Deep
29